\pdfoutput=1

\documentclass[11pt]{article}

\usepackage{EMNLP2022}

\usepackage{times}
\usepackage{latexsym}
\usepackage{subfig}
\usepackage{graphicx}
\usepackage{amsmath}
\usepackage{mathrsfs}
\usepackage{amsmath,amsthm,amsfonts,amssymb,amscd}
\usepackage{soul}
\usepackage{url}
\usepackage{multicol,multirow}
\usepackage{tabularx}

\usepackage[T1]{fontenc}

\usepackage[utf8]{inputenc}

\usepackage{microtype}

\usepackage{inconsolata}

\title{Legal and Political Stance Detection of SCOTUS Language}

\author{Noah Bergam \\
Columbia University \\
New York, NY \\
  \texttt{njb2154@columbia.edu} 
  \\\And
  Emily Allaway \\
  Columbia University \\
    New York, NY \\
  \texttt{eallaway@cs.columbia.edu} 
  \\\And
  Kathleen McKeown \\
  Columbia University \\
    New York, NY \\
  \texttt{kathy@cs.columbia.edu} 
  \\}

\begin{document}
\maketitle
\begin{abstract}
We analyze publicly available US Supreme Court documents using automated stance detection. In the first phase of our work, we investigate the extent to which the Court's public-facing language is political. We propose and calculate two distinct ideology metrics of SCOTUS justices using oral argument transcripts. We then compare these language-based metrics to existing social scientific measures of the ideology of the Supreme Court and the public. Through this cross-disciplinary analysis, we find that justices who are more responsive to public opinion tend to express their ideology during oral arguments. This observation provides a new kind of evidence in favor of the attitudinal change hypothesis of Supreme Court justice behavior. As a natural extension of this political stance detection, we propose the more specialized task of \textit{legal} stance detection with our new dataset \texttt{SC-stance}, which matches written opinions to legal questions. We find competitive performance on this dataset using language adapters trained on legal documents. 





\end{abstract}

\section{Introduction}

The relationship between the Supreme Court of the United States (SCOTUS) and American public opinion is complicated. Some scholars  debate normative questions as to whether the Court's power of judicial review ought to obey democratic principles\footnote{Famously dubbed the "counter-majoritarian difficulty" by political scientist Alexander Bickel in 1962, this problem has been said to lie at the heart of American Constitutional scholarship \cite{friedman1998history}} \cite{bassok2013solving}. Others investigate how SCOTUS behaves in relation to the public and why \cite{katz2017general} . 
Prior work in the field of American political science has consistently demonstrated an association between the partisan ideology of the Court, as expressed through its decisions, and that of the public, as recorded through poll data \citep[e.g.,]{casillas2011public,mishler1996public}.
However, more recent work, particularly in light of the 2022 Dobbs v. Jackson decision, suggests a departure from this general pattern  ~\cite{jessee2022decade}. This change in institutional behavior has profound social significance which calls for academic attention.
This paper heeds that call by providing a new analytical perspective on SCOTUS' democratic tendencies.

Despite extensive research confirming SCOTUS' general responsiveness to public opinion, the underlying reasoning for this relationship is disputed. One hypothesis centers on \textit{strategic behavior}: 
it posits that the Court consciously acts in accordance with the public will in order to protect its Constitutionally fragile claim to the power of judicial review~\citep{hammond2005strategic}. Alternatively, the \textit{attitudinal change hypothesis} contends that broader socio-political forces such as news media present confounding factors that influence both the justices and the public~\citep{norpoth1994popular}. 

In this paper, we gain new insight into these hypotheses by applying automated stance detection to a newly assembled corpus of Supreme Court written opinions and oral arguments. Stance detection (i.e., automatically identifying the position of an author towards a given target statement) allows us to evaluate the implications 
of a justice's language. We use stance detection and related techniques to build two different textual indicators of ideology which we call \textit{issue-specific stance} (ISS) and \textit{holistic political stance} (HPS) respectively. We compare these indicators to existing social scientific metrics related to general public opinion (i.e. the Stimson Policy Mood; \citealp{stimson2018public}), Supreme Court justice ideology (i.e. the Martin-Quinn score; \citealp{martin2002dynamic}), and Supreme Court case salience (i.e. the Clark case salience; \citealp{clark2015measuring}).

\begin{table*}[t]
  \centering
  \resizebox{\textwidth}{!}{
  \begin{tabular}{ | p{0.05\linewidth}p{0.92\linewidth}| }
  \hline
  $D_1 = $
  &  \textit{Once the Court starts looking to the currents of public opinion regarding a particular judgment, it enters a truly bottomless pit from which there is simply no extracting itself.} (Rehnquist, 1992) \\\hline
    $D_2 = $
    & \textit{Will this institution survive the stench that this creates in the public perception that the Constitution and its reading are just political acts?} (Sotomayor, 2022)\\\hline
    \(T=\) & \textit{The Supreme Court ought to make decisions with the public opinion in mind.} \\ \hline
    &\hspace{2.5cm} \textbf{stance($D_1, T$) = con} \hspace{2cm}
\textbf{stance($D_2, T$) = pro} \\ \hline
  \end{tabular}
  }
  \caption{A relevant, sophisticated example of stance detection.}
  \label{tab:1}
\end{table*}


In addition, we build a supervised stance detection dataset,
\texttt{SC-stance}, over a subset of Supreme Court written opinions. 
Our dataset matches the text of the written opinion to a corresponding legal question (i.e., the target) posed on a legal educational website\footnote{\url{Oyez.org}}. We present baselines on this dataset using tf-idf features, 
two language models for the legal domain~\cite{chalkidis2020legal,zheng2021does}, and a new method which involves augmenting BERT~\cite{devlin2018bert} with an adapter~\cite{pfeiffer2020adapterhub} pre-trained for the legal domain. We find performance gains 
both 
with this new method
and from masking named entities in the training data.



The main contributions of this work are
as follows. 
\textbf{(1)} Using stance detection, we formulate two distinct ideology metrics (i.e. \textit{holistic political stance} and \textit{issue-specific stance}) for SCOTUS justices serving from 1955 to 2020. We find that justices who are responsive to public opinion tend to use language which correlates ideologically with their voting behavior. This provides new evidence in favor of the attitudinal change hypothesis. \textbf{(2)} We release a new dataset, \texttt{SC-stance}, which matches written opinion text to relevant legal questions. It is the first \textit{legal stance detection} dataset as far as the authors are aware. 
\textbf{(3)} We set baselines on our new dataset and find two ways to potentially improve performance: 
using a 
law-specific language adapter, and removing named entities during training.



The repository of relevant code is publicly available through the following link: 
\href{https://github.com/njbergam/scotus-public-stance}{https://github.com/njbergam/scotus-public-stance}.

\section{Related Work}

\paragraph{Supreme Court and Public Opinion}
There is extensive academic work analyzing the Supreme Court’s relationship with public opinion. In some cases, facts about the Supreme Court are gauged using a public opinion-related proxy. For instance, \citet{segal1989ideological} developed an ideology score of justices based on newspaper editorials written at the time of their appointment while \citet{epstein2000measuring} and \citet{clark2015measuring} used front-page news articles in order to quantify the political salience of Supreme Court cases. Other projects take a more direct look at the correlation between SCOTUS decisions and public opinion metrics. \citet{casillas2011public} uses a two-step least-squares regression approach in order to trace the public’s influence on Court voting patterns, while \citet{kastellec2010public} looks at the relationship between state-level public opinion polls and Senator votes for SCOTUS justice nominations. 



A common thread in many prior studies is the focus on Court voting behavior or its reception in the public eye. In contrast, our work investigates how SCOTUS presents its politics through its \textit{language}. This approach takes advantage of the fact that the corpus of official SCOTUS language is publicly available, relatively small, and well-structured. 

Previous work in various fields demonstrates that there are concrete differences between the language used by people of different political ideologies. In psycholinguistics, \citet{robinson2017mind} suggests that the language of liberals tends to emphasize mental concepts, while that of conservatives uses more references to the body. NLP research has further investigated this concept
through political ideology detection on two datasets \citep[e.g.,]{iyyer2014political}:
Convote (i.e. Congressional dialogue labeled with the political affiliation of the speaker) \cite{thomas-etal-2006-get}, and the Ideological Books Corpus (i.e. sentences from political articles and books annotated for political cues) \cite{sim2013measuring}.

\paragraph{Legal Artifical Intelligence}
The legal domain presents a unique challenge for NLP due to the precision, structure, and everyday importance of legal language~\cite{dale2019law}. Furthermore, legal language is interesting in terms of its intersection with political discourse\footnote{This intersection can be problematic. The Code of Conduct for US Judges states: "A Judge Should Refrain from Political Activity" \cite{uscourtsCodeConduct} and presents restrictions on language, e.g. no public endorsement of political candidates.}, a much more well-studied genre in NLP. In this work, we investigate that very intersection by leveraging existing stance and political ideology detection datasets in the context of legal language.  

There are two major types of legal AI models~\cite{zhong2020does}: rule-based methods, which are mostly supported by legal AI practitioners in industry, and embedding-based methods, which seem to garner the most attention from researchers in academia. 
The latter body of work has recently focused on adapting pre-trained language models (e.g., BERT) to the legal domain,
either through law-specific pre-training, fine-tuning, or a combination of both \cite{chalkidis2020legal, zheng2021does}. Due to the general accessibility of many legal documents around the world, a wide variety of legal NLP datasets are now available, six of which were recently consolidated into the LexGLUE benchmark~\cite{chalkidis2021lexglue}. 
Our dataset, \texttt{SC-stance}, provides a test of legal understanding which is not currently captured by existing datasets. Rather than evaluating the relevance between legal statements or documents, \texttt{SC-stance} goes a step further and tests the relative stance. 


\paragraph{Stance Detection}
The task of stance detection is to determine the stance (e.g., \texttt{Pro}, \texttt{Con}, or \texttt{Neutral}) of a text on a target (e.g., `abortion')~\cite{Mohammad2016SemEval2016T6}
(see Table 1 for an illustration). 
In many works on stance detection, the topic is a noun-phrase (e.g., `legalization of abortion') and texts are relatively short, such as posts from debate forums~\cite[e.g.,]{Abbott2016InternetAC,Walker2012ACF,Hasan2014WhyAY}, and comments on news articles~\cite{Krejzl2017StanceDI,allaway-mckeown-2020-zero}.
Stance detection on Twitter towards political targets is particularly popular~\cite{sobhani-etal-2017-dataset,Li2021PStanceAL,Cignarella2020SardiStanceE,Lai2020MultilingualSD,Taul2017OverviewOT}.
Despite this interest, there is a lack of labeled stance data in the legal domain. Our dataset \texttt{SC-stance} not only fills this gap, it also challenges stance detection systems with complex targets (i.e., full sentences) and long documents (i.e., thousands of words).




\begin{table}[t]
\begin{center}
\begin{tabular}{ lll|rr} 
 \hline
  \textbf{Metric} & \textbf{Dataset} & \textbf{Model} & $F1$ & \textbf{Acc. }\\  \hline
 \multirow{2}{*}{ISS} & \multirow{2}{*}{VAST} & Baseline & 58.2 & - \\
 & & Ours & 62.8 & 63.4\\
 \hline
 \multirow{2}{*}{HPS} & \multirow{2}{*}{Convote} & Baseline & - & 70.2\\
 & & Ours & 75.3 & 76.3\\
 \hline
\end{tabular}
\caption{
Performance of the stance detection classifiers. The baseline for VAST is a BERT-based model~\cite{allaway-mckeown-2020-zero} and for Convote it is an RNN~\cite{iyyer2014political}. 
}
\end{center}
\end{table}

\section{Evaluating Political Stance}

\subsection{Methods}\label{sec:analysismethods}

In the first phase of our work, we track how Supreme Court justices express political leanings in their public-facing language. We focus on two particular corpora: the set of written opinions (1789-2020), and the set of oral argument transcripts (1955-2020). The former was obtained through a Kaggle database \cite{fiddler} which used the Harvard CaseLaw Project's\footnote{\href{https://case.law/}{https://case.law/}} API to collect full text files of $33,490$ Supreme Court written opinions.
The latter was scraped from the Oyez Project~\cite{urofsky2001oyez}, a multimedia archive of SCOTUS data. We collected over \(3.8\) million lines of dialogue

\subsubsection{Linguistic Ideology Metrics}
Stance detection allows us to represent the political polarity of judicial language through our two new ideology metrics: issue-specific stance (ISS) and holistic political stance (HPS). Both measure a speaker's ideology along the classic liberal-conservative spectrum \cite{stimson2012meaning}. However, they arrive at their answers very differently. The ISS evaluates a speaker's stance relative to a set of representative topics, while the HPS seeks to classify the political affiliation of the speaker directly.  Both metrics are built on top of transformer-based text classification algorithms. Although the ISS and HPS are calculated by statement, it is understood that each requires large representative samples of a speaker's statements in to provide some insight into their overall ideology.

\paragraph{Issue-specific stance (ISS)} To obtain a speaker's ISS, we gauge a speaker’s stance on various liberal and conservative political statements. We adapt these statements from the Pew Political Typology Quiz \cite{pewresearchPoliticalTypology}, which uses a variety of questions to evaluate ideology on a continuous scale from "Progressive Left" to "Faith and Flag Conservative." Based on how much the given text agrees or disagrees with each of the liberal and conservative statements (which are paraphrased for simplicity), we construct a score which is meant to gauge ideology.

If a higher score indicates a conservative leaning (this is, of course, an arbitrary choice), then we can frame the ISS calculation for a specific text \(t\) as follows. Given a set of targets which align roughly with liberal ideals \(S_L\), a conservative counterpart \(S_C\), and a stance model which maps to some signed interval \([-1,1]\) we calculate ISS as follows:
\[\text{ISS}_{S_L, S_C}(t) = \sum_{l \in S_C} s(l, t) - \sum_{c \in S_L} s(c, t)\]
We formulate the above stance model as giving a continuous output. In practice, this amounts to adding the softmax probability of the predicted class, signed according to the ideology of the statement. 


\paragraph{Holistic political stance (HPS)} This metric 
seeks to immediately classify whether a given piece of language expresses more conservative or liberal ideology overall. As such, the underlying detector is not trained to detect stance relative to a specific topic; rather, it is trained to predict the ideology of the speaker. 
This framework may help provide a
a broader psychological perspective on the underlying ideology of someone's language. For instance, suppose \citet{robinson2017mind} is correct that liberals and conservatives generally express metaphors differently. Then HPS may pick up on that implicit ideological cue if it noticed such a pattern in its training data. In contrast, ISS is, by design, better at picking up on explicit cues such as the affirmation of a liberal or conservative belief. Additionally, HPS is simple to calculate (i.e., it is the confidence output of a binary ideology classifier). Unlike ISS, it does not require the parameters of liberal and conservative targets. This inherent simplicity also makes the \(HPS\) algorithm run faster.

\subsubsection{Calculating HPS and ISS}

ISS and HPS rely on pre-trained stance and ideology classification models, respectively. This means they require different datasets for training. For the ISS metric, we train a model using the Varied Stance Topics (VAST) dataset~\cite{allaway-mckeown-2020-zero}, which covers a large range of mostly political topics with broad themes like climate change and immigration. For the HPS metric, we train a classifier using the Convote dataset~\cite{Thomas+Pang+Lee:06a}, which maps statements spoken by Congressional representatives to their partisan affiliation. We formulate both of these as binary stance classification tasks for the sake of simplicity.  

Our classifiers use RoBERTa~\cite{liu2019roberta} without fine-tuning.  As shown in Table 2, we achieve higher accuracy than the existing baselines in the original papers for each datset

To obtain the \(\text{ISS}\) or \(\text{HPS}\) score of a justice in a particular time period, we first collect the set of statements which contain some sort of emotion, with the intuition that this would increase the likelihood that the statement contains an opinion (as opposed to boilerplate legal language). To do this, we collect statements which feature a word from the NRC Emotion Lexicon \cite{mohammad2013nrc}. Then, for each justice, we collect a representative sample of statements per year
and take the average score over all of these statements, to get the HPS of that particular year. In our experiments in the next section, we took sample sizes on the order of \(10^3\) per year per judge due to the time constraints of processing the text.  

\subsubsection{Baseline Metrics}
We compare our linguistic ideology metrics to 
three existing metrics in the quantitative political sciences. These will serve as important baselines 
since
they help us contextualize and evaluate our own metrics. These metrics have been calculated in previous research and are available through online databases.

\textbf{Martin-Quinn scores}\footnote{\href{https://mqscores.lsa.umich.edu/}{mqscores.lsa.umich.edu/}} are dynamic ideal-point estimations of justices’ political ideologies~\cite{martin2002dynamic}. This metric, calculated on a yearly basis, uses a latent variable model where a justice's voting behavior is the observed variable.

\textbf{The Stimson Policy Mood} \footnote{\href{https://stimson.web.unc.edu/data/}{stimson.web.unc.edu/data/}} gauges the general political leanings of the public through longitudinal surveys, which ask questions on a variety of issues over repeated time points \cite{stimson2018public}.

\textbf{The Clark case salience}\footnote{\href{https://dataverse.harvard.edu/dataset.xhtml?persistentId=doi:10.7910/DVN/29637}{dataverse.harvard.edu/dataset.xhtml...}} metric uses front page newspaper articles in The New York Times, The Washington Post, and The L.A. Times to quantify how relevant different Supreme Court cases are in the public eye \cite{clark2015measuring}.

\subsection{Results}
The first round of our analysis centers on the relationship between our linguistic ideology metrics and existing measures of Supreme Court behavior. 


\begin{figure}[h]
\centering
\includegraphics[width=0.38\textwidth]{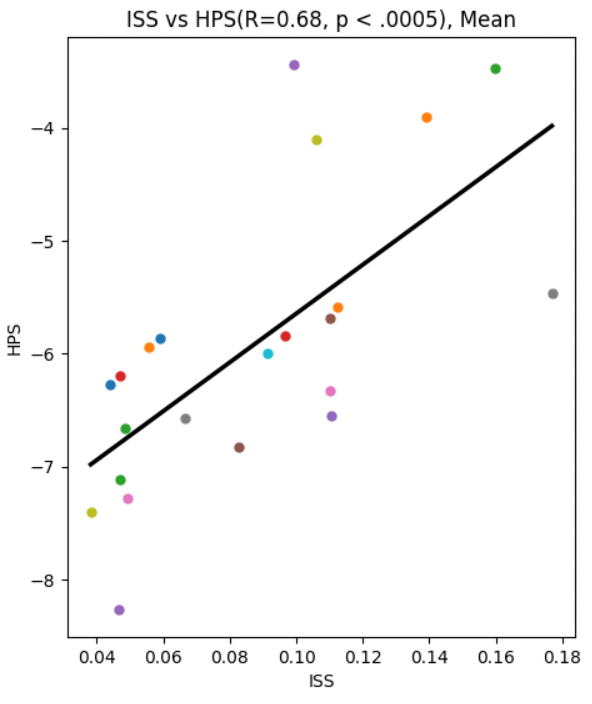}
\caption{A strong correlation (\(R=0.68, p<0.0005\)) between the holistic and the issue-specific stance scores. Each data point represents a justice's mean score over their tenure (the significance drops to \(p<0.0001\) when we consider their median score).}
\label{fig:ISS_HPS}
\end{figure}

\paragraph{ISS and HPS correlate.} We first undertake a simple methodological audit and compare the issue-specific and holistic political stance scores across 23 justices who served from \(1955\) to \(2020\). We find that the two correlate quite strongly (Fig~\ref{fig:ISS_HPS}), despite the fact that the underlying stance detectors were formulated and trained in very different ways (\S\ref{sec:analysismethods}). This suggests that the detectors are measuring the same signal and provides evidence that there is in fact a political signal in the dialogues of justices. This is not only important for our analysis, but it is also surprising in its own right given the officially apolitical stance of the Supreme Court \cite{uscourtsCodeConduct}.

\begin{figure*}[h!]
    \centering
    \subfloat
    {{\includegraphics[width=0.4\textwidth]{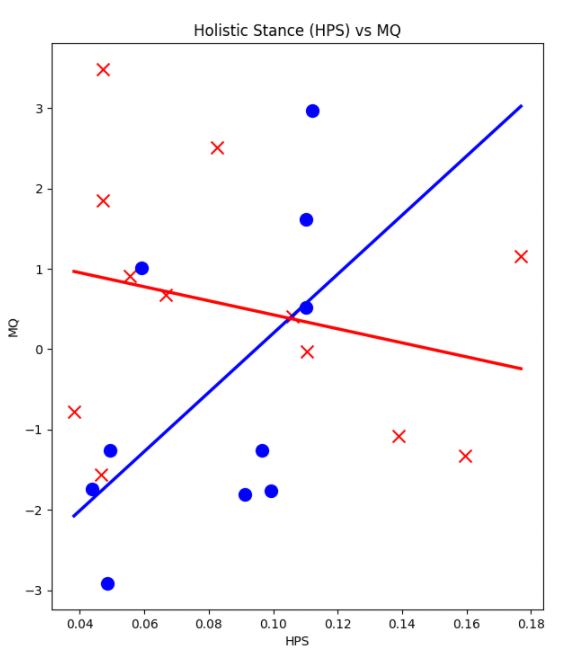} }}%
    \subfloat
    {{\includegraphics[width=0.524\textwidth]{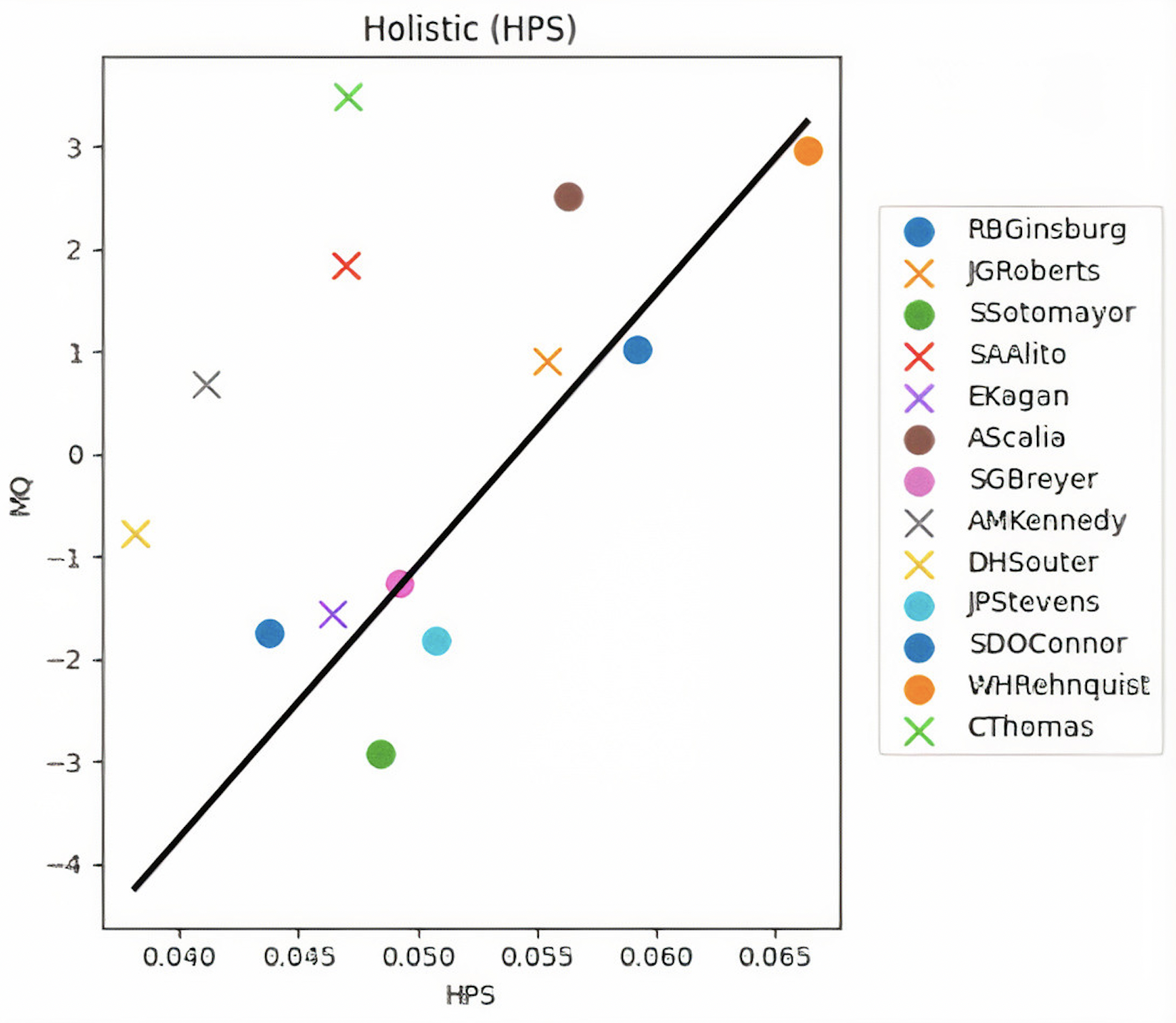} }}%
    \caption{Mean holistic political stance versus mean Martin-Quinn score, by justice. In both figures, circles represent justices whose MQ scores correlate significantly with the Stimson policy mood over their tenure as justices.  The left graph shows justices from 1955 to 2020, while the right shows and labels justices only after 1990. HPS was obtained using random sampling, with \(n=2000\) statements per year in the left graph and \(n=1000\) statements per year on the right.}%
    \label{fig:MQscores}%
\end{figure*}

\paragraph{Insight on the Attitudinal Change Hypothesis.} 
Next, we looked at our metrics (ISS and HPS) in relation to the Martin-Quinn score. Importantly, we partition the justices based on their general responsiveness to public opinion. We measure this responsiveness by gauging the correlation between yearly Martin-Quinn scores (i.e., estimating justices' ideology) and the Stimson policy mood (i.e., estimating public opinion), by justice. We say that justices are "responsive" if this correlation is significant with \(p < 0.05\).

We found that justices who are more responsive to the public opinion, compared to their counterparts, exhibit a much greater correlation between the ideology of their language, as measured by ISS and HPS and that of their voting decisions (Fig~\ref{fig:MQscores}). This pattern is particularly noticeable with the HPS score. Additionally, this pattern intensifies when we looked purely at justices who have served past \(1990\).

This result offers new support for the \textit{attitudinal change hypothesis}, which explains the correlation between Supreme Court decisions and public opinion 
by arguing that ``the same social forces that shape 
the mass public also influence Supreme Court justices''~\cite{casillas2011public}. 

Our results support the \textit{attitudinal change hypothesis} for two reasons. 
Firstly, note that a major underlying assumption of attitudinal change is that ``individual attitudes are assumed to be the primary determinants of behavior''~\cite{mishler1996public}. Thus, if justices are responsive to public opinion because of their attitudes, then these attitudes would affect both voting behavior and language. This is precisely what we observe when we find a correlation between Martin-Quinn scores and HPS for responsive justices.

Furthermore, the \textit{strategic behavior hypothesis} does not have as much explanatory power for our results. HPS, by design, is sensitive to speech patterns that mirror those of Congresspeople. Considering the norms of the Court, it is more likely that such quasi-political behavior stems from latent, ideological influences rather than strategic behavior. If anything, strategic behavior would explain a correlation between ISS (i.e. explicit ideological expression) and MQ, which we did not observe.

\begin{figure}[h]
\centering
\resizebox{\columnwidth}{!}{
\includegraphics[width=0.5\textwidth]{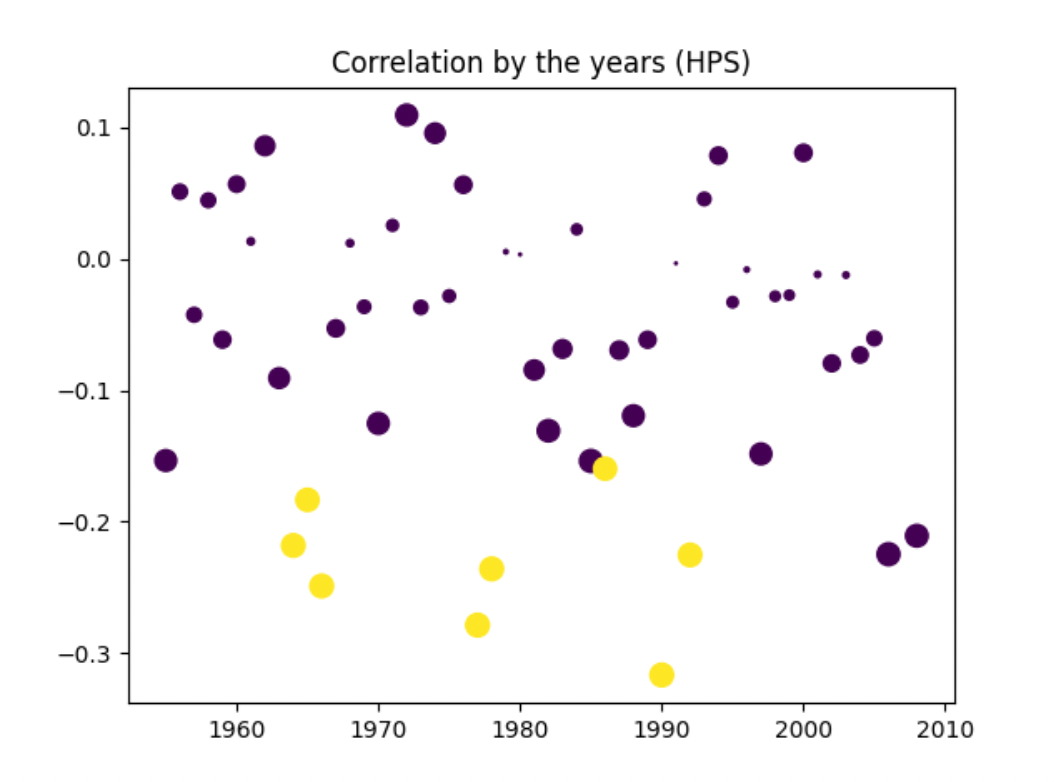}
}
\caption{Correlation between confidence of the HPS score and the Clark case salience over all Supreme Court written opinions from 1955 to 2008. Yellow denotes a statistically significant correlation.}
\label{fig:clarkcase}
\end{figure}

\paragraph{Case salience and political language.} 

We also consider political undertones of written opinions. We analyzed the relationship between the \textit{magnitude} of the HPS of the written opinion text (a measure of its general political signal) and the Clark Case salience (i.e., public relevance) of the corresponding case. We found that the correlation was almost always slightly negative and only statistically significant for a handful of years (Fig~\ref{fig:clarkcase}). 


This seemingly negative result actually parallels previous findings. In particular, \citet{casillas2011public} argue that public opinion may (counter-intuitively) hold less of an influence on salient cases as opposed to non-salient cases, since non-salient cases are simply more frequent. If the use of political language in a ruling can be seen as response to public opinion –– which would seem to be the case under either of the leading hypotheses of Supreme Court behavior –– then our result supports the theory of an inverse relationship between salience and politicality.





\section{\texttt{SC-stance} dataset}

\subsection{Methods}
We describe the collection and characteristics of our new stance dataset, \texttt{SC-stance}, as well as the methods we apply to it.

\begin{table}[h]
\begin{center}
\resizebox{\columnwidth}{!}{
\begin{tabular}{ |p{0.475\textwidth}| } 
\hline
 \textbf{Case}: School District of Abington Township v. Schempp (ID 1962-148), Majority Opinion. \\ \hline
\textbf{Target}: Did the Pennsylvania law requiring public school students to participate in classroom religious exercises violate the religious freedom of students as protected by the First and Fourteenth Amendments? \\ \hline
\textbf{Text}: 
Once again we are called upon to consider the scope of the provision of the First Amendment to the United States Constitution which declares that "Congress shall make no law respecting an establishment of religion, or prohibiting the free exercise thereof" [...] In light of the history of the First Amendment and of our cases interpreting and applying its requirements, \hl{we hold that the practices at issue and the laws requiring them are unconstitutional under the Establishment Clause, as applied to the States through the Fourteenth Amendment}. [...] \\ \hline
\textbf{Label}: \texttt{pro} (text affirms the target) \\
 \hline
\end{tabular}}
\caption{An example data point from \texttt{SC-stance}, in which we highlight the relevant portion of the text which confirms the stance.}
\label{tab:dataexample}
\end{center}
\end{table}

Our dataset \texttt{SC-stance} was drawn from three sources: a dataset of full-text Supreme Court opinions through 2020~\cite{fiddler}, the Washington University Supreme Court Database~\cite{spaeth2014supreme}, and the Oyez website~\cite{urofsky2001oyez}. We started by collecting written opinions which had non-neutral holdings, as encoded in the SC Database. We then automatically matched these opinion texts to the key legal question on the Oyez website to obtain text-target pairs.
Since the questions on Oyez are always phrased such that an affirmative answer is in favor of the petitioner, we used the Winning Party label\footnote{\href{http://scdb.wustl.edu/documentation.php?var=partyWinning}{scdb.wustl.edu/documentation.php?var=partyWinning}} 
 from the Supreme Court Database, as well as the opinion type given in the Kaggle dataset (i.e. majority, concurring, dissenting, etc.) to infer the stance that a given written opinion takes towards the legal question (e.g. if the winning party was the respondent, and the opinion type was dissenting, then the opinion affirms the legal question).
 
The final dataset has $2708$ labeled instances ($1179$ labeled \texttt{pro}, $930$ labeled \texttt{con}). The average length of a target (i.e., the legal question) is $35$ tokens and the average length of a text (i.e., the Supreme Court written opinion) is $5330$ tokens. We show an example datapoint in Table~\ref{tab:dataexample}.


In addition to providing a legal stance detection task, our dataset could provide an interesting passage retrieval task. Most other legal information retrieval datasets map documents to other documents (e.g., the German Dataset for Legal Information Retrieval~\cite{wrzalik2021gerdalir}) or to static questions which are unchanged between documents (e.g., the Contract Understanding Atticus Dataset~\cite{hendrycks2021cuad}). The closest counterpart to our dataset, to the best of our knowledge, is the Belgian Statutory Article Retrieval Dataset, a French language dataset that maps legal questions written by laypeople to Belgian law articles~\cite{louis2021statutory}. 

\subsubsection{Models for Stance Detection}

\begin{figure}[t]
\centering
\resizebox{\columnwidth}{!}{
\includegraphics[width=0.5\textwidth]{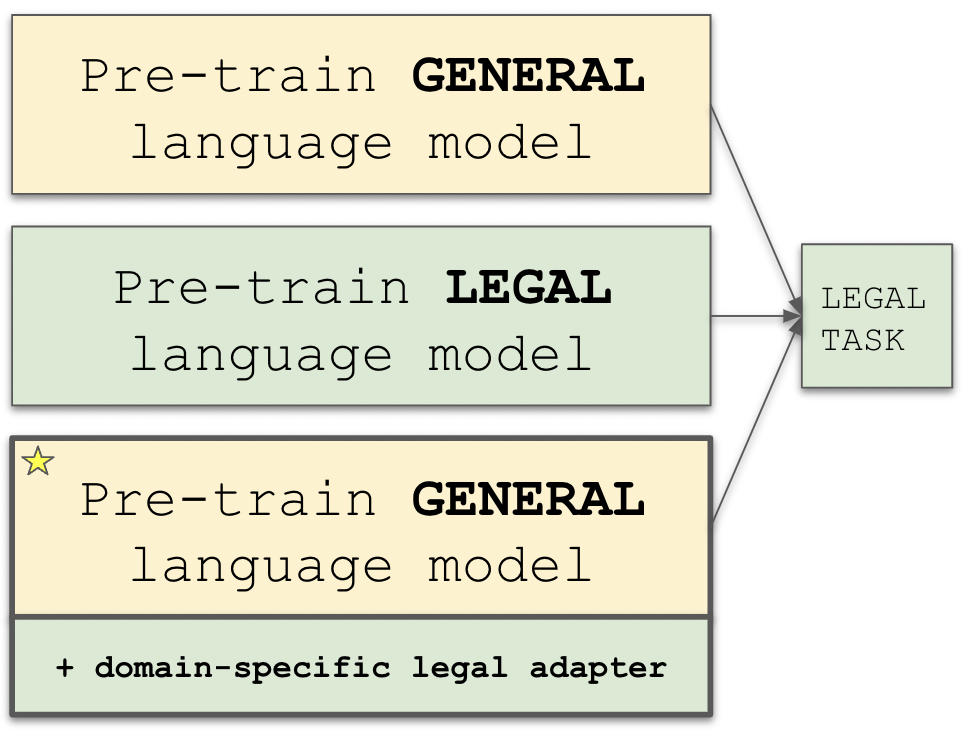}
}
\caption{Three methods of tackling a legal NLP task using a large language model (the third being our new method which leverages language adapters). This paradigm generalizes to other domain-specific applications such as medicine or finance. }
\label{fig:models}
\end{figure}

In comparing models, we are most interested in which ones learn the most informative \textit{features} from the text. The final layer is, in almost all cases, a single layer feed-forward network (Fig~\ref{fig:models}).

\paragraph{Legal Adapter} Inspired by the concept that ``legalese'' could potentially be treated as a unique language, we use a \textit{language adapter} to transfer a BERT-based stance detection model from its training data's domain to the \texttt{SC-stance} dataset. It is important to note that Supreme Court opinion language is relatively clear and concise compared to the more pure legalese of contracts or securities filings. While it may seem conceptually extreme to treat SCOTUS filings as a separate language, it is experimentally interesting as it sheds light on whether a dedicated adaptation for legal language allows for a more effective automated reading of legal stance.

 Adapters have been used to enable efficient multilingual transfer for language models. An adapter module is a set of weights (i.e., feed-forward layers) inserted into each attention block of a transformer and trained using masked language modeling (MLM). Adapters were originally designed as an alternative to fine-tuning~\cite{houlsby2019parameter} and have since become a popular method of cross-lingual domain transfer~\cite[e.g.]{pfeiffer2020mad, vidoni2020orthogonal}. 
One intuitive benefit of this approach over pre-training an entire language model is that only unlabeled data is needed to train the adapter and training is more parameter efficient, since the adapter has comparatively few parameters.



\begin{table*}[t!]
\centering
{
\begin{tabularx}{\textwidth}{X|rr|rr}
\hline
 & \multicolumn{2}{>{\centering\arraybackslash}X|}{\textbf{Binary}}
 & \multicolumn{2}{>{\centering\arraybackslash}X}{
 \textbf{3-class}}\\
  & \textbf{Original}  & \textbf{w/ NER-mask} & \textbf{Original} & \textbf{w/ NER-mask} \\ \hline\hline
 Majority & 39.6  & - & 20.4 & - \\ \hline
 tf-idf (LR) & 41.4 & 43.2 & 26.5 &  29.6 \\
 tf-idf (MLP) & 50.0  & 49.8 & 32.0 & 31.5  \\ \hline
 BERT & 50.4 & 47.1 & 36.9 & 35.1 \\ 
 CaseLaw-BERT & 47.6 & 49.2 & 38.3 & 40.3 \\ 
 Legal-BERT & 52.8 & 53.0 & \textbf{47.4} & 41.7 \\ \hline
 Legal Adapter & \textbf{55.6} & \textbf{53.4} & 41.4 & \textbf{42.2} \\ 
 \hline
\end{tabularx}}
\caption{F-1 scores on the \texttt{SCS-written} dataset, using an 80-20 train-test split. 
}
\label{tab:results}
\end{table*}

\paragraph{Baselines} We compare our new method to a number of baselines, the simplest being the tf-idf vectorization of each of the target and document. On these simple features, we compare logistic regression (LR) and multilayer perceptron (MLP) as final layers; we find that the latter performs significantly better with \(p < 0.02\)\footnote{We use an approximate randomization test.}, so we proceed to use MLP as the classification layer in our BERT-based models.

We experiment with BERT \cite{devlin2018bert}, a popular transformer-based encoder pre-trained with masked language modeling and next sentence prediction. We also investigate two variants, which differ largely in terms of their training corpus. One is Legal-BERT \cite{chalkidis2020legal}, which is pre-trained on an English legal corpus and uses a sub-word vocabulary built from scratch. The other is CaseLaw-BERT \cite{zheng2021does}, which is pre-trained on the Harvard Law case corpus.


\subsubsection{Experimental Details}
We evaluated our stance models on \texttt{SC-stance} in two settings: binary classification (i.e., labels \{pro, con\}) and 3-class classification (i.e., \{pro, con, neutral\}). Since \texttt{SC-stance} does not have any neutral labeled instances, following \citet{allaway-mckeown-2020-zero} we randomly pair opinions with unrelated questions to augment the dataset. For the adapter, we follow ~\citet{pfeiffer2020adapterhub} and train a  legal language adapter using MLM for $230k$ epochs with a learning rate of \(10^{-4}\) and a batch size of $16$. As unlabeled data we use over 8.8 million sentences from case law documents, made available through SigmaLaw~\cite{osfSigmaLawLarge}. In all experiments the \texttt{SC-stance} dataset is split 80/20 for training and testing. Importantly, we consider the case in which the training set has all named entities (with the notable exception of laws) masked during the training phase\footnote{We masked named entities using the Python spacy library's 'en\_core\_web\_sm' model. The mask was the named entity type: for instance, "October 10" would become "[DATE] [DATE]".} and revealed during testing. This is referred to as the \textit{NER-mask} setting in Table~\ref{tab:results}. For BERT and its variants, we append the legal question followed by a '\texttt{[SEP]}' token and the written opinion, and we truncate past the 512 token limit, with the understanding that most written opinions, despite their length, express their stance early on.

\subsection{Results}




Overall, we found that the legal adapter is competitive with the leading legal language models, achieving the highest F-1 score (\(55.6\)) on the binary classification task\footnote{Due to the small size of the dataset, we were unable to mark these differences as statistically significant.}. In the 3-class setting, it was only outperformed by Legal-BERT. 

We found that Legal-BERT consistently outperforms BERT and CaseLaw-BERT (\(p < 0.09\) for the 3-class setting), which corroborates 
the experiments of Legal-BERT's creators~\cite{chalkidis2020legal}.
We also found that, while the BERT-based features consistently outperformed the ``classical'' counterparts, the tf-idf model with an MLP classification layer had strong performance on the binary classification task.  

We found mixed results with the NER mask setting, in that it led to both gradual increases (e.g. tf-idf with logistic regression, CaseLaw-BERT) as well as considerable drops in performance (e.g. legal adapter binary, Legal-BERT the 3-class setting). Intuitively, the NER mask should remove spurious signals for the classifier, since the relationship between the target and topic should almost never be related to the entities (i.e. proper nouns), but instead the relationships between entities. 

We believe this hypothetical advantage is what led to certain score increases. However, the flip-side is that there may be instability introduced when the model is presented with proper nouns in the test setting, after having had them removed during training. We noticed that BERT was more susceptible to this instability, which may be attributable to its less specialized vocabulary or understanding of legal grammar. These weaknesses of domain shift may increase the model's susceptibility to spurious signals.


\section{Conclusion}

Using state-of-the-art NLP techniques, we gain new insight into a longstanding political science problem: the Supreme Court's relationship with public opinion. In our analysis of the language of Supreme Court justices, we leverage existing metrics of SCOTUS behavior as well as stance detection datasets regarding political ideology. Notably, we find a new source of evidence for the attitudinal change hypothesis of the Supreme Court, and we experiment with a competitive new model for legal language domain adaptation. 


This research sheds light on how stance detection allows us to interrogate the implicit opinions of static documents. This is a powerful use case of NLP for the social sciences, in that it allows for a large-scale, critical analysis of large bodies of text. Of course, there is a long way to go in the field of stance detection, both generally and in specific linguistic domains such as the law. Our contribution of \texttt{SC-stance} feeds into this goal, by providing semantically rich targets and a mix of legal and lay language. We emphasize this latter feature, in that quality textual understanding -- for human and AI alike -- is marked by a thorough comprehension of both colloquial and technical language formulation.


\section*{Limitations}

Our stance detection analysis of Supreme Court language is a proof-of-concept experiment with considerable potential for expansion. For instance, one could obtain a much richer understanding of Supreme Court ideology using a flavor of stance detection which analyzes targets relevant to issues of \textit{jurisprudence} (e.g. judicial activism, originalism) rather than common politics.  

There is also room for expansion in terms of our use and formulation of certain metrics. For instance, we chose not to investigate ``public opinion'' through text data, partly because the concept has no clear-cut representative corpus, and sampling from the web or the news could present selection biases. However, this problem could be resolved with a narrower view of public opinion such as, say, the news media. The inherent benefit to having a text-based metric of public opinion is that it is more easily comparable to text-based metrics of Supreme Court ideology.  Furthermore, it may be enlightening to track the partisanship of justices’ language using ideal point estimation (i.e., the words are the observed variable, the ideology is the hidden variable), rather than direct measurement of the justice stance year after year \cite{bafumi2005practical}.

In terms of processing the \texttt{SC-stance} dataset, future work should look into how to work with the long written opinions using BERT-based methods which have a token limit. There is also clear potential to expand the \texttt{SC-stance} dataset. This could be done through strategic web-scraping of certioriari petitions, which often contain the relevant legal questions of (what eventually becomes) a Supreme Court case. If this challenge of locating the petitions, scraping the relevant text, and matching to the relevant case can be met, then the \texttt{SC-stance} dataset could in principle grow by orders of magnitude, which would make it an even more promising ground for experimentation. 


\section*{Ethics Statement}

Our investigation of the Supreme Court is an academic exploration of a political subject. By employing stance detection, we mean to uncover large-scale patterns in the text which may not be obvious to a single reader or scholar. This should not take away from the pursuit of engaging with text directly. After all, by transforming text into statistics, we lose many dimensions of its complexity in order to zero in on specific attributes. It is important to acknowledge this methodological complexity as quantitative social sciences research continues to engage with NLP-driven metadata.

\section*{Acknowledgements}

The authors thank Hayley Cohen, Suresh Naidu,  and the anonymous reviewers at NLLP for their comments. We thank Ariella Lang for organizing the Laidlaw Scholars Program which provided funding and support for this project. We also recognize the Columbia NLP Group for providing computing resources which made these experiments possible. 



\bibliography{my}

\begin{thebibliography}{53}
\expandafter\ifx\csname natexlab\endcsname\relax\def\natexlab#1{#1}\fi

\bibitem[{Abbott et~al.(2016)Abbott, Ecker, Anand, and
  Walker}]{Abbott2016InternetAC}
Rob Abbott, Brian Ecker, Pranav Anand, and Marilyn~A. Walker. 2016.
\newblock Internet argument corpus 2.0: An sql schema for dialogic social media
  and the corpora to go with it.
\newblock In \emph{LREC}.

\bibitem[{Allaway and McKeown(2020)}]{allaway-mckeown-2020-zero}
Emily Allaway and Kathleen McKeown. 2020.
\newblock \href {https://doi.org/10.18653/v1/2020.emnlp-main.717}
  {{Z}ero-{S}hot {S}tance {D}etection: {A} {D}ataset and {M}odel using
  {G}eneralized {T}opic {R}epresentations}.
\newblock In \emph{Proceedings of the 2020 Conference on Empirical Methods in
  Natural Language Processing (EMNLP)}, pages 8913--8931, Online. Association
  for Computational Linguistics.

\bibitem[{Bafumi et~al.(2005)Bafumi, Gelman, Park, and
  Kaplan}]{bafumi2005practical}
Joseph Bafumi, Andrew Gelman, David~K Park, and Noah Kaplan. 2005.
\newblock Practical issues in implementing and understanding bayesian ideal
  point estimation.
\newblock \emph{Political Analysis}, 13(2):171--187.

\bibitem[{Bassok and Dotan(2013)}]{bassok2013solving}
Or~Bassok and Yoav Dotan. 2013.
\newblock Solving the countermajoritarian difficulty?
\newblock \emph{International journal of constitutional law}, 11(1):13--33.

\bibitem[{Casillas et~al.(2011)Casillas, Enns, and
  Wohlfarth}]{casillas2011public}
Christopher~J Casillas, Peter~K Enns, and Patrick~C Wohlfarth. 2011.
\newblock How public opinion constrains the us supreme court.
\newblock \emph{American Journal of Political Science}, 55(1):74--88.

\bibitem[{Center(2021)}]{pewresearchPoliticalTypology}
Pew~Research Center. 2021.
\newblock {P}olitical {T}ypology {Q}uiz --- pewresearch.org.
\newblock \url{https://www.pewresearch.org/politics/quiz/political-typology/}.

\bibitem[{Chalkidis et~al.(2020)Chalkidis, Fergadiotis, Malakasiotis, Aletras,
  and Androutsopoulos}]{chalkidis2020legal}
Ilias Chalkidis, Manos Fergadiotis, Prodromos Malakasiotis, Nikolaos Aletras,
  and Ion Androutsopoulos. 2020.
\newblock Legal-bert: The muppets straight out of law school.
\newblock \emph{arXiv preprint arXiv:2010.02559}.

\bibitem[{Chalkidis et~al.(2021)Chalkidis, Jana, Hartung, Bommarito,
  Androutsopoulos, Katz, and Aletras}]{chalkidis2021lexglue}
Ilias Chalkidis, Abhik Jana, Dirk Hartung, Michael Bommarito, Ion
  Androutsopoulos, Daniel~Martin Katz, and Nikolaos Aletras. 2021.
\newblock Lexglue: A benchmark dataset for legal language understanding in
  english.
\newblock \emph{arXiv preprint arXiv:2110.00976}.

\bibitem[{Cignarella et~al.(2020)Cignarella, Lai, Bosco, Patti, and
  Rosso}]{Cignarella2020SardiStanceE}
Alessandra~Teresa Cignarella, Mirko Lai, Cristina Bosco, Viviana Patti, and
  Paolo Rosso. 2020.
\newblock \href{http://ceur-ws.org/Vol-2765/paper159.pdf}{SardiStance @
  EVALITA2020: Overview of the Task on Stance Detection in Italian Tweets}.
\newblock In \emph{EVALITA}.

\bibitem[{Clark et~al.(2015)Clark, Lax, and Rice}]{clark2015measuring}
Tom~S Clark, Jeffrey~R Lax, and Douglas Rice. 2015.
\newblock Measuring the political salience of supreme court cases.
\newblock \emph{Journal of Law and Courts}, 3(1):37--65.

\bibitem[{Courts(2019)}]{uscourtsCodeConduct}
United~States Courts. 2019.
\newblock {C}ode of {C}onduct for {U}nited {S}tates {J}udges --- uscourts.gov.
\newblock
  \url{https://www.uscourts.gov/judges-judgeships/code-conduct-united-states-judges#f}.
\newblock [Accessed 01-Oct-2022].

\bibitem[{Dale(2019)}]{dale2019law}
Robert Dale. 2019.
\newblock Law and word order: Nlp in legal tech.
\newblock \emph{Natural Language Engineering}, 25(1):211--217.

\bibitem[{de~Silva(2019)}]{osfSigmaLawLarge}
Nisansa de~Silva. 2019.
\newblock {S}igma{L}aw - {L}arge {L}egal {T}ext {C}orpus and {W}ord
  {E}mbeddings --- osf.io.
\newblock \url{https://osf.io/qvg8s/}.

\bibitem[{Devlin et~al.(2018)Devlin, Chang, Lee, and
  Toutanova}]{devlin2018bert}
Jacob Devlin, Ming-Wei Chang, Kenton Lee, and Kristina Toutanova. 2018.
\newblock Bert: Pre-training of deep bidirectional transformers for language
  understanding.
\newblock \emph{arXiv preprint arXiv:1810.04805}.

\bibitem[{Epstein and Segal(2000)}]{epstein2000measuring}
Lee Epstein and Jeffrey~A Segal. 2000.
\newblock Measuring issue salience.
\newblock \emph{American Journal of Political Science}, pages 66--83.

\bibitem[{Fiddler(2020)}]{fiddler}
Garrett Fiddler. 2020.
\newblock Scotus opinions.
\newblock Full text and metadata of all opinions written by SCOTUS justices
  through 2020,
  \url{https://www.kaggle.com/datasets/gqfiddler/scotus-opinions}.

\bibitem[{Friedman(1998)}]{friedman1998history}
Barry Friedman. 1998.
\newblock The history of the countermajoritarian difficulty, part one: The road
  to judicial supremacy.
\newblock \emph{NYUL Rev.}, 73:333.

\bibitem[{Hammond et~al.(2005)Hammond, Bonneau, and
  Sheehan}]{hammond2005strategic}
Thomas~H Hammond, Chris~W Bonneau, and Reginald~S Sheehan. 2005.
\newblock \emph{Strategic behavior and policy choice on the US Supreme Court}.
\newblock Stanford University Press.

\bibitem[{Hasan and Ng(2014)}]{Hasan2014WhyAY}
Kazi~Saidul Hasan and Vincent Ng. 2014.
\newblock Why are you taking this stance? identifying and classifying reasons
  in ideological debates.
\newblock In \emph{EMNLP}.

\bibitem[{Hendrycks et~al.(2021)Hendrycks, Burns, Chen, and
  Ball}]{hendrycks2021cuad}
Dan Hendrycks, Collin Burns, Anya Chen, and Spencer Ball. 2021.
\newblock Cuad: An expert-annotated nlp dataset for legal contract review.
\newblock \emph{arXiv preprint arXiv:2103.06268}.

\bibitem[{Houlsby et~al.(2019)Houlsby, Giurgiu, Jastrzebski, Morrone,
  De~Laroussilhe, Gesmundo, Attariyan, and Gelly}]{houlsby2019parameter}
Neil Houlsby, Andrei Giurgiu, Stanislaw Jastrzebski, Bruna Morrone, Quentin
  De~Laroussilhe, Andrea Gesmundo, Mona Attariyan, and Sylvain Gelly. 2019.
\newblock Parameter-efficient transfer learning for nlp.
\newblock In \emph{International Conference on Machine Learning}, pages
  2790--2799. PMLR.

\bibitem[{Iyyer et~al.(2014)Iyyer, Enns, Boyd-Graber, and
  Resnik}]{iyyer2014political}
Mohit Iyyer, Peter Enns, Jordan Boyd-Graber, and Philip Resnik. 2014.
\newblock Political ideology detection using recursive neural networks.
\newblock In \emph{Proceedings of the 52nd Annual Meeting of the Association
  for Computational Linguistics (Volume 1: Long Papers)}, pages 1113--1122.

\bibitem[{Jessee et~al.(2022)Jessee, Malhotra, and Sen}]{jessee2022decade}
Stephen Jessee, Neil Malhotra, and Maya Sen. 2022.
\newblock A decade-long longitudinal survey shows that the supreme court is now
  much more conservative than the public.
\newblock \emph{Proceedings of the National Academy of Sciences},
  119(24):e2120284119.

\bibitem[{Kastellec et~al.(2010)Kastellec, Lax, and
  Phillips}]{kastellec2010public}
Jonathan~P Kastellec, Jeffrey~R Lax, and Justin~H Phillips. 2010.
\newblock Public opinion and senate confirmation of supreme court nominees.
\newblock \emph{The Journal of Politics}, 72(3):767--784.

\bibitem[{Katz et~al.(2017)Katz, Bommarito, and Blackman}]{katz2017general}
Daniel~Martin Katz, Michael~J Bommarito, and Josh Blackman. 2017.
\newblock A general approach for predicting the behavior of the supreme court
  of the united states.
\newblock \emph{PloS one}, 12(4):e0174698.

\bibitem[{Krejzl et~al.(2017)Krejzl, Hourov{\'a}, and
  Steinberger}]{Krejzl2017StanceDI}
Peter Krejzl, Barbora Hourov{\'a}, and Josef Steinberger. 2017.
\newblock \href{https://dl.acm.org/doi/abs/10.1145/3369026}{Stance detection in
  online discussions}.
\newblock \emph{ArXiv}, abs/1701.00504.

\bibitem[{Lai et~al.(2020)Lai, Cignarella, Far{\'i}as, Bosco, Patti, and
  Rosso}]{Lai2020MultilingualSD}
Mirko Lai, Alessandra~Teresa Cignarella, D.~I.~H. Far{\'i}as, Cristina Bosco,
  V.~Patti, and P.~Rosso. 2020.
\newblock
  \href{https://www.sciencedirect.com/science/article/pii/S0885230820300085}{Multilingual
  stance detection in social media political debates}.
\newblock \emph{Comput. Speech Lang.}, 63:101075.

\bibitem[{Li et~al.(2021)Li, Sosea, Sawant, Nair, Inkpen, and
  Caragea}]{Li2021PStanceAL}
Yingjie Li, Tiberiu Sosea, Aditya Sawant, Ajith~Jayaraman Nair, Diana Inkpen,
  and Cornelia Caragea. 2021.
\newblock \href{https://aclanthology.org/2021.findings-acl.208/}{P-Stance: A
  Large Dataset for Stance Detection in Political Domain}.
\newblock In \emph{FINDINGS}.

\bibitem[{Liu et~al.(2019)Liu, Ott, Goyal, Du, Joshi, Chen, Levy, Lewis,
  Zettlemoyer, and Stoyanov}]{liu2019roberta}
Yinhan Liu, Myle Ott, Naman Goyal, Jingfei Du, Mandar Joshi, Danqi Chen, Omer
  Levy, Mike Lewis, Luke Zettlemoyer, and Veselin Stoyanov. 2019.
\newblock Roberta: A robustly optimized bert pretraining approach.
\newblock \emph{arXiv preprint arXiv:1907.11692}.

\bibitem[{Louis et~al.(2021)Louis, Spanakis, and
  Van~Dijck}]{louis2021statutory}
Antoine Louis, Gerasimos Spanakis, and Gijs Van~Dijck. 2021.
\newblock A statutory article retrieval dataset in french.
\newblock \emph{arXiv preprint arXiv:2108.11792}.

\bibitem[{Martin and Quinn(2002)}]{martin2002dynamic}
Andrew~D Martin and Kevin~M Quinn. 2002.
\newblock Dynamic ideal point estimation via markov chain monte carlo for the
  us supreme court, 1953--1999.
\newblock \emph{Political analysis}, 10(2):134--153.

\bibitem[{Mishler and Sheehan(1996)}]{mishler1996public}
William Mishler and Reginald~S Sheehan. 1996.
\newblock Public opinion, the attitudinal model, and supreme court decision
  making: A micro-analytic perspective.
\newblock \emph{The Journal of Politics}, 58(1):169--200.

\bibitem[{Mohammad et~al.(2016)Mohammad, Kiritchenko, Sobhani, Zhu, and
  Cherry}]{Mohammad2016SemEval2016T6}
Saif~M. Mohammad, Svetlana Kiritchenko, Parinaz Sobhani, Xiao-Dan Zhu, and
  Colin Cherry. 2016.
\newblock Semeval-2016 task 6: Detecting stance in tweets.
\newblock In \emph{SemEval@NAACL-HLT}.

\bibitem[{Mohammad and Turney(2013)}]{mohammad2013nrc}
Saif~M Mohammad and Peter~D Turney. 2013.
\newblock Nrc emotion lexicon.
\newblock \emph{National Research Council, Canada}, 2:234.

\bibitem[{Norpoth et~al.(1994)Norpoth, Segal, Mishler, and
  Sheehan}]{norpoth1994popular}
Helmut Norpoth, Jeffrey~A Segal, William Mishler, and Reginald~S Sheehan. 1994.
\newblock Popular influence on supreme court decisions.
\newblock \emph{American Political Science Review}, 88(3):711--724.

\bibitem[{Pfeiffer et~al.(2020{\natexlab{a}})Pfeiffer, R{\"u}ckl{\'e}, Poth,
  Kamath, Vuli{\'c}, Ruder, Cho, and Gurevych}]{pfeiffer2020adapterhub}
Jonas Pfeiffer, Andreas R{\"u}ckl{\'e}, Clifton Poth, Aishwarya Kamath, Ivan
  Vuli{\'c}, Sebastian Ruder, Kyunghyun Cho, and Iryna Gurevych.
  2020{\natexlab{a}}.
\newblock Adapterhub: A framework for adapting transformers.
\newblock \emph{arXiv preprint arXiv:2007.07779}.

\bibitem[{Pfeiffer et~al.(2020{\natexlab{b}})Pfeiffer, Vuli{\'c}, Gurevych, and
  Ruder}]{pfeiffer2020mad}
Jonas Pfeiffer, Ivan Vuli{\'c}, Iryna Gurevych, and Sebastian Ruder.
  2020{\natexlab{b}}.
\newblock Mad-x: An adapter-based framework for multi-task cross-lingual
  transfer.
\newblock \emph{arXiv preprint arXiv:2005.00052}.

\bibitem[{Robinson et~al.(2017)Robinson, Boyd, Fetterman, and
  Persich}]{robinson2017mind}
Michael~D Robinson, Ryan~L Boyd, Adam~K Fetterman, and Michelle~R Persich.
  2017.
\newblock The mind versus the body in political (and nonpolitical) discourse:
  Linguistic evidence for an ideological signature in us politics.
\newblock \emph{Journal of Language and Social Psychology}, 36(4):438--461.

\bibitem[{Segal and Cover(1989)}]{segal1989ideological}
Jeffrey~A Segal and Albert~D Cover. 1989.
\newblock Ideological values and the votes of us supreme court justices.
\newblock \emph{American Political Science Review}, 83(2):557--565.

\bibitem[{Sim et~al.(2013)Sim, Acree, Gross, and Smith}]{sim2013measuring}
Yanchuan Sim, Brice~DL Acree, Justin~H Gross, and Noah~A Smith. 2013.
\newblock Measuring ideological proportions in political speeches.
\newblock In \emph{Proceedings of the 2013 conference on empirical methods in
  natural language processing}, pages 91--101.

\bibitem[{Sobhani et~al.(2017)Sobhani, Inkpen, and
  Zhu}]{sobhani-etal-2017-dataset}
Parinaz Sobhani, Diana Inkpen, and Xiaodan Zhu. 2017.
\newblock \href {https://aclanthology.org/E17-2088} {A dataset for multi-target
  stance detection}.
\newblock In \emph{Proceedings of the 15th Conference of the {E}uropean Chapter
  of the Association for Computational Linguistics: Volume 2, Short Papers},
  pages 551--557, Valencia, Spain. Association for Computational Linguistics.

\bibitem[{Spaeth et~al.(2014)Spaeth, Epstein, Ruger, Whittington, Segal, and
  Martin}]{spaeth2014supreme}
Harold Spaeth, Lee Epstein, Ted Ruger, Keith Whittington, Jeffrey Segal, and
  Andrew~D Martin. 2014.
\newblock Supreme court database code book.
\newblock \emph{URL: http://scdb. wustl. edu}.

\bibitem[{Stimson(2012)}]{stimson2012meaning}
James~A Stimson. 2012.
\newblock On the meaning \& measurement of mood.
\newblock \emph{Daedalus}, 141(4):23--34.

\bibitem[{Stimson(2018)}]{stimson2018public}
James~A Stimson. 2018.
\newblock \emph{Public opinion in America: Moods, cycles, and swings}.
\newblock Routledge.

\bibitem[{Taul{\'e} et~al.(2017)Taul{\'e}, Mart{\'i}, Pardo, Rosso, Bosco, and
  Patti}]{Taul2017OverviewOT}
Mariona Taul{\'e}, Maria~Ant{\`o}nia Mart{\'i}, Francisco M.~Rangel Pardo,
  Paolo Rosso, Cristina Bosco, and Viviana Patti. 2017.
\newblock
  \href{https://iris.unito.it/retrieve/handle/2318/1652739/371876/Overview5.pdf}{Overview
  of the Task on Stance and Gender Detection in Tweets on Catalan
  Independence}.
\newblock In \emph{IberEval@SEPLN}.

\bibitem[{Thomas et~al.(2006{\natexlab{a}})Thomas, Pang, and
  Lee}]{thomas-etal-2006-get}
Matt Thomas, Bo~Pang, and Lillian Lee. 2006{\natexlab{a}}.
\newblock \href {https://aclanthology.org/W06-1639} {Get out the vote:
  Determining support or opposition from congressional floor-debate
  transcripts}.
\newblock In \emph{Proceedings of the 2006 Conference on Empirical Methods in
  Natural Language Processing}, pages 327--335, Sydney, Australia. Association
  for Computational Linguistics.

\bibitem[{Thomas et~al.(2006{\natexlab{b}})Thomas, Pang, and
  Lee}]{Thomas+Pang+Lee:06a}
Matt Thomas, Bo~Pang, and Lillian Lee. 2006{\natexlab{b}}.
\newblock Get out the vote: Determining support or opposition from
  {Congressional} floor-debate transcripts.
\newblock In \emph{Proceedings of EMNLP}, pages 327--335.

\bibitem[{Urofsky(2001)}]{urofsky2001oyez}
Melvin~I Urofsky. 2001.
\newblock The oyez project: Us supreme court multimedia database.
\newblock \emph{The Journal of American History}, 88(2):753.

\bibitem[{Vidoni et~al.(2020)Vidoni, Vuli{\'c}, and
  Glava{\v{s}}}]{vidoni2020orthogonal}
Marko Vidoni, Ivan Vuli{\'c}, and Goran Glava{\v{s}}. 2020.
\newblock Orthogonal language and task adapters in zero-shot cross-lingual
  transfer.
\newblock \emph{arXiv preprint arXiv:2012.06460}.

\bibitem[{Walker et~al.(2012)Walker, Tree, Anand, Abbott, and
  King}]{Walker2012ACF}
Marilyn~A. Walker, Jean E.~Fox Tree, Pranav Anand, Rob Abbott, and Joseph King.
  2012.
\newblock A corpus for research on deliberation and debate.
\newblock In \emph{LREC}.

\bibitem[{Wrzalik and Krechel(2021)}]{wrzalik2021gerdalir}
Marco Wrzalik and Dirk Krechel. 2021.
\newblock Gerdalir: A german dataset for legal information retrieval.
\newblock In \emph{Proceedings of the Natural Legal Language Processing
  Workshop 2021}, pages 123--128.

\bibitem[{Zheng et~al.(2021)Zheng, Guha, Anderson, Henderson, and
  Ho}]{zheng2021does}
Lucia Zheng, Neel Guha, Brandon~R Anderson, Peter Henderson, and Daniel~E Ho.
  2021.
\newblock When does pretraining help? assessing self-supervised learning for
  law and the casehold dataset of 53,000+ legal holdings.
\newblock In \emph{Proceedings of the Eighteenth International Conference on
  Artificial Intelligence and Law}, pages 159--168.

\bibitem[{Zhong et~al.(2020)Zhong, Xiao, Tu, Zhang, Liu, and
  Sun}]{zhong2020does}
Haoxi Zhong, Chaojun Xiao, Cunchao Tu, Tianyang Zhang, Zhiyuan Liu, and Maosong
  Sun. 2020.
\newblock How does nlp benefit legal system: A summary of legal artificial
  intelligence.
\newblock \emph{arXiv preprint arXiv:2004.12158}.

\end{thebibliography}
\bibliographystyle{acl_natbib}

\end{document}